\documentclass[manuscript,screen]{acmart}
\usepackage{subcaption}\newcounter{example}[section]
\newenvironment{example}[1][]{\refstepcounter{example}\par\medskip
   \noindent \textbf{Example~\theexample. #1} \rmfamily}{\medskip}

\usepackage{todonotes}

\begin{document}

\title[Multi-dimensional discrimination in Law and Machine Learning - A comparative overview]
{Multi-dimensional discrimination in Law and Machine Learning - A comparative overview}




\author{Arjun Roy}
\authornote{Both authors contributed equally to this research.}
\email{roy@l3s.de}
\orcid{0000-0002-4279-9442}
\affiliation{%
  \institution{L3S Research Center Hannover; Institute of Computer Science, Free University of Berlin}
  \country{Germany}
}
\author{Jan Horstmann}
\authornotemark[1]
\email{jan.horstmann@iri.uni-hannover.de}
\affiliation{%
  \institution{Institute for Legal Informatics, Leibniz University of Hanover}
  \country{Germany}
}
\author{Eirini Ntoutsi}
\affiliation{%
  \institution{Research Institute CODE, Bundeswehr University Munich}
  \country{Germany}
}

\begin{abstract}
AI-driven decision-making can lead to discrimination against certain individuals or social groups based on protected characteristics/attributes such as race, gender, or age. The domain of fairness-aware machine learning focuses on methods and algorithms for understanding, mitigating, and accounting for bias in AI/ML models. Still, thus far, the vast majority of the proposed methods assess fairness based on a single protected attribute, e.g. only gender or race. In reality, though, human identities are multi-dimensional, and discrimination can occur based on more than one protected characteristic, leading to the so-called ``multi-dimensional discrimination'' or ``multi-dimensional fairness'' problem. While well-elaborated in legal literature, the multi-dimensionality of discrimination is less explored in the machine learning community. Recent approaches in this direction mainly follow the so-called intersectional fairness definition from the legal domain, whereas other notions like additive and sequential discrimination are less studied or not considered thus far. In this work, we overview the different definitions of multi-dimensional discrimination/fairness in the legal domain as well as how they have been transferred/ operationalized (if) in the fairness-aware machine learning domain. By juxtaposing these two domains, we draw the connections, identify the limitations, and point out  open research directions.
\end{abstract}




\maketitle
\section{Introduction}
\label{sec:intro}
AI-driven decision-making has already penetrated into almost all spheres of human life, from content recommendation~\cite{liu2020fairrec} and healthcare~\cite{panesar2019AIhealth} to predictive policing~\cite{alikhademi2021AIpolicing} and autonomous driving~\cite{vidano2023autodrive}, deeply affecting everyone, anywhere, anytime. In a variety of cases, discriminatory impacts of AI-driven decision-making on individuals or social groups on the basis of the \emph{protected attributes} like gender, race, age, and others have been observed. Examples range from recidivism prediction~\cite{dressel2018accuracy}, hiring~\cite{raghavan2020mitigating}, recommendations~\cite{pitoura2021fairness} to healthcare~\cite{cahan2019putting}, education~\cite{hu2020towards}, service provision~\cite{jackson2018algorithmic} and  surveillance~\cite{urquhart2022policing}. Discriminatory impacts concern both symbolic or representative equality (e.g. ads related to arrest records appearing more frequently along search results for names associated with blacks than whites~\cite{sweeneyaddelivery}) and distributive equality (regarding the access to social goods, e.g. in biased hiring algorithms)~\cite{xenidissenden}.
The domain of fairness-aware machine learning~\cite{pedreshi2008discrimination}, is concerned with bias and discrimination in AI systems and covers a wide range of topics from understanding bias and discrimination to methods for bias mitigation and accountability~\cite{ntoutsi2020bias}. Fair ML research has also been taken up in legal scholarship, with debate as to what role statistical fairness metrics can play under anti-discrimination law~\cite{hackerfairness, whyfairnesscannot, wachterbiaspreservation, HAUER2021105583, Gentgenbarg2022}.
However, despite this steadily growing body of research, the vast majority of proposed methods assumes that discrimination is based on a single protected attribute, for example, only gender or only race. We refer to this as \emph{mono-dimensional discrimination/fairness} or \emph{mono-discrimination/fairness}. 
For the mono-discrimination case, several fairness definitions have been proposed, see~\cite{verma2018fairness} for a survey, as well as methods for mitigating mono-discrimination, e.g.~\cite{kamiran2010discrimination,krasanakis2018adaptive,iosifidis2019adafair}.


In reality though, humans have \emph{multi-dimensional} identities~\cite{uccellarimultiple2008}. We all have a gender, racial or ethnic origin, age and sexual orientation (and more), and are \emph{categorised} by others according to such concepts~\cite{baerbittnergoettsche2010mehrdimensional}. Consequently, discrimination cannot always be attributed to a single protected attribute but rather many protected attributes, for example, a \emph{combination} of gender, race \emph{and} age can be the basis of discrimination, leading to the problem of \emph{multi-dimensional discrimination/fairness} or \emph{multi-discrimination/fairness}~\footnote{We use the terms multi-discrimination and multi-fairness interchangeably.}. Empirical evidence consistently suggests that discrimination in the real world is often multi-dimensional~\cite{FRAopinion2022}. E.g. in a 2015 Eurobarometer, approximately a quarter of discrimination cases, as reported by the persons affected, was based on multiple grounds~\cite{eurobarometer2015discrimination}. Multi-dimensional discrimination especially impacts ethnic minorities, as evidenced by a finding from 2010 by the European Fundamental Rights Agency (FRA) that 14\% of respondents from ethnic minorities indicated feeling discriminated against on multiple grounds in the 12 months prior to the survey, with most multi-dimensional discrimination experienced by ethnic minority women~\cite{FRAreport2010}. Completing the picture, scholars in the field also assume that a large share of discrimination is multi-dimensional~\cite{xenidis2020_EU_equalitylaw} or even consider multi-dimensional discrimination the rule and mono-dimensional discrimination the exception~\cite{baerbittnergoettsche2010mehrdimensional}. Multi-dimensional discrimination also seems to play an important role in AI systems, as concrete examples of, e.g. low accuracy of facial analysis for black females~\cite{pmlr-v81-buolamwini18a} indicate. 
However, only in the last years the topic of multi-discrimination has caught the attention of the fairness-aware ML community~\cite{kearns2018gerrymandering,pmlr-multicalibration,foulds2020intersectional,foulds2020bayesian-intersectional,yang2020counterfactual-intersectional,yang2020fairness_overlap,kang2021multifair_MI,roy2022multifair}.
Considering multi-dimensional discrimination in ML algorithms introduces new challenges, from how to define fairness in the presence of multiple protected attributes to how to mitigate multi-discrimination. For the former, we find that the concept of intersectionality from the legal domain is mainly adapted, whereas other concepts like cumulative and sequential discrimination are less used or developed. 
For the latter, a key challenge for ML is data scarcity as protected subgroups defined by the intersection of multiple protected attributes become smaller or even empty as the number of protected attributes increases.

The goal of this survey is to draw attention to this important topic and provide an overview of existing approaches in legal and ML literature. We intend to thereby contribute to a ``legal-technical argumentation framework''~\cite{Legal-technical}. Juxtaposing types of multi-dimensional discrimination from legal scholarship with operational definitions of multi-fairness in ML enables us to draw connections between the two domains, highlight differences on the conceptual level and identify directions for future research.
The remainder of the paper is organised as follows: We start with an introduction of multi-dimensionality in law and a typology of multi-dimensional discrimination (Section~\ref{sec:multiDefsLaw}) with concrete examples. Definitions of multi-dimensional discrimination in ML and associated challenges are discussed in Section~\ref{sec:multiDefsML}.
A critical comparison of the two domains is presented in Section~\ref{sec:disc} followed by open challenges and directions for future work in Section~\ref{sec:outlook}.

\section{Multi-dimensional discrimination in Law}
\label{sec:multiDefsLaw}
While we introduce and illustrate the concept of multi-dimensional discrimination from legal scholarship with a view to current EU private anti-discrimination law\footnote{This field is mainly codified in four EU Directives: Directive 2000/43/EC of 29 June 2000 implementing the principle of equal treatment between persons irrespective of racial or ethnic origin; Directive 2000/78/EC of 27 November 2000 establishing a general framework for equal treatment in employment and occupation; Directive 2004/113/EC of 13 December 2004 implementing the principle of equal treatment between men and women in the access to and supply of goods and services; Directive 2006/54/EC of 5 July 2006 on the implementation of the principle of equal opportunities and equal treatment of men and women in matters of employment and occupation (recast).} and the exact situation in other legal systems is beyond our scope, the concept of multi-dimensionality can be applied to all anti-discrimination provisions. Discrimination consists of either a \emph{direct} discriminatory treatment or, in the case of \emph{indirect} discrimination, the application of a seemingly neutral criterion, provision or practice leading to a particular disadvantage of a group or individual based on a \emph{protected ground}\footnote{While the terms ground (law) and protected attribute (ML literature) both refer to the criterion, e.g. sex, to be protected - against discriminatory treatment/impact or bias -, a 1-1 mapping between these two terms cannot be assumed. This is discussed in \ref{sec:categoricalgrounds}.}.
EU private anti-discrimination law, like many other systems, is applicable only to exhaustively listed grounds of discrimination\footnote{To that effect for Directive 2000/78/EC, ECJ, C-13/05 - \emph{Chacón Navas} par. 56; C-306/06 - \emph{Coleman} par. 46 and C-354/13 - \emph{Kaltoft} par. 36.\label{ECJ analogy}}, which are \emph{racial and ethnic origin}, \emph{sex}, \emph{religion and belief}, \emph{disability}, \emph{age} and \emph{sexual orientation}. 
Although the issue of interactions between different discrimination grounds has been pointed out by scholarship, this development has largely been met with hesitance in legal practice. EU law rather vaguely recognises ``multiple'' discrimination of women\footnote{Mentioned, but not elaborated on, in recital 14 of Directive 2000/43/EC and recital 3 of Directive 2000/78/EC. Note that EU member state law may differ: of note, Spain passed a law explicitly covering intersectional discrimination in 2022, Ley 15/2022, de 12 de julio, integral para la igualdad de trato y la no discriminación.\label{fnEUlawmultipletext}}. Despite this textual reference, the European Court of Justice (ECJ) has cautiously evaded the issue and on one occasion expressed resistance to the notion of intersectional discrimination as a special type of discrimination~\footnote{Cf., as the latest examples C-808/18 - \emph{WABE eV} par. 58 and case C-443/15 - \emph{Parris} par. 80, for analysis see~\cite{howard2018CJEU}.}. Hence, our discussion in this regard is mostly limited to scholarship of EU law.
The consideration of the interaction of different grounds of discrimination in legal literature started with the coining of the term ``intersectionality'' in 1989~\cite{crenshaw1989intersectional}. Critical scholarship has since developed a multitude of concepts to better analyse the reality of discrimination. There is no universally accepted terminology for discrimination involving more than one protected ground~\cite{fredman2016intersectional}.
Due to the lack of a legally defined term, we choose to speak of the \emph{multi-dimensionality of discrimination} as has been suggested in literature~\cite{hutchinson2001multidimensional,baerbittnergoettsche2010mehrdimensional,schiek2005multidimensional,schiekmultiintorduction}. This term is sufficiently broad to capture a variety of \emph{interactions among different protected attributes}, from one incident of discrimination involving multiple grounds to multiple discriminatory incidents over time.
The ways in which different grounds interact can be used to distinguish further between different types of discrimination. We use a common typology~\cite{uccellarimultiple2008,fredman2016intersectional}, which identifies three types of interactions, cumulative or additive discrimination (Section~\ref{sec:cumulativeLaw}), intersectional discrimination (Section~\ref{sec:intersectionalLaw}) and sequential discrimination (Section~\ref{sec:sequentialLaw}).

\subsection{Cumulative discrimination}
\label{sec:cumulativeLaw}
In cumulative (often also termed additive) discrimination, a disadvantage is linked to two or more grounds of discrimination, e.g. gender and race. These are, however, \emph{conceptually separable}, meaning that one can identify distinct disadvantages linked to each involved ground which ``add up'' when the grounds are observed together. 

\begin{figure}[httb]
    \centering
         \centering
\includegraphics[scale=0.6]{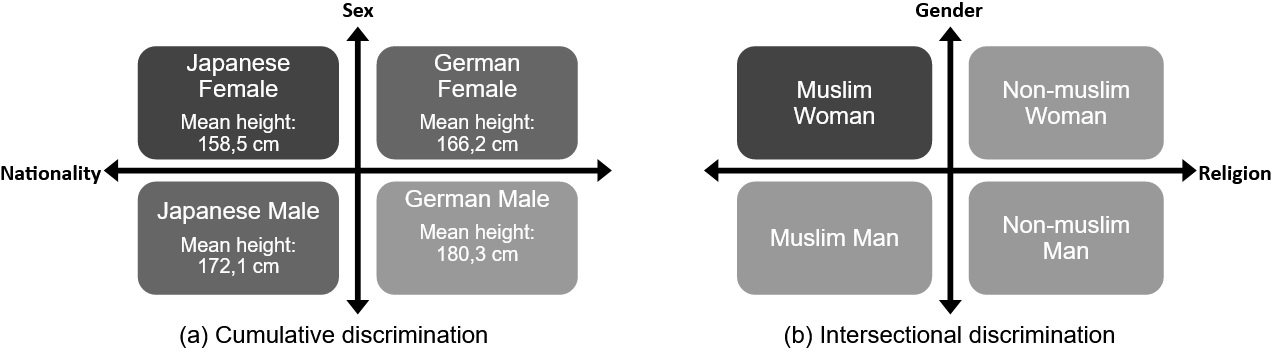}
    \caption{(a) Cumulative discrimination: the impact of height requirements according to nationality and sex; 
    (b) intersectional discrimination: the prohibition of headscarves specifically affects Muslim women. 
    Darker shades indicate stronger impact. Sex refers to the biological property influencing height, gender to social roles influencing the wearing of a headscarf according to religious practices.}
    \label{fig:boxplot_pums}
\end{figure}

\begin{example}
A hypothetical example of cumulative  discrimination is shown in Figure~\ref{fig:boxplot_pums}(a), displaying the mean height of four subgroups defined on the basis of the two grounds sex and nationality. Height varies statistically between sexes and nationalities (here, we use mean height of 19-year olds in 2019 according to~\cite{meanheight}).
The impact of height requirements, e.g. for jobs in the security sector~\cite{noack_2018}, thus differs depending on nationality (which in turn correlates with ethnic origin) and sex, but the disadvantage ``adds up'' for women of certain nationality (and thus, ethnic origin).
\end{example}

\subsection{Intersectional discrimination}
\label{sec:intersectionalLaw}

In intersectional discrimination, the grounds of discrimination involved are merged into one and \emph{cannot be separated} in  the analysis. Intersectional discrimination thus affects subgroups defined by a combination of grounds. 

\begin{example}
A concrete example of intersectional discrimination is shown in Figure~\ref{fig:boxplot_pums} (b). The prohibition of wearing headscarves (as has been discussed and implemented for teachers in some kindergartens in Germany\footnote{See, e.g. ECJ, C-804/18 - \emph{WABE eV}. Such rules usually concern ``visible religious symbols'', leading to the issue whether this constitutes direct or indirect discrimination~\cite{howardheadscarves}.}) specifically affects religious Muslim women, a subgroup of both women and Muslims.
\end{example}

\subsection{Sequential discrimination}
\label{sec:sequentialLaw}
In sequential discrimination, discrimination occurs on the basis of the same or different grounds over several incidents in temporal sequence (see Example~\ref{ex:sequential}). 

\begin{figure}[httb]
    \centering
    \includegraphics[scale=0.6]{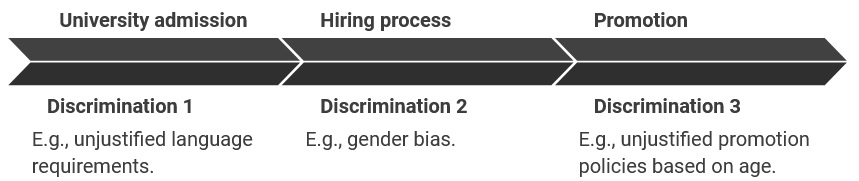}
    \caption{An example of sequential discrimination in work life based on three different grounds.}
    \label{fig:sequential}
\end{figure}

\begin{example}
An example can be found in Figure~\ref{fig:sequential}, where potential points of discrimination in a person's work life are shown. Discrimination at the earlier stages is likely to also affect the outcome at later stages.
\label{ex:sequential}
\end{example}

This is the ``normal case'' for anti-discrimination law. To quote~\cite{fredman2016intersectional}: ``This is perhaps the easiest to deal with. Each incident can be assessed on a single ground and compensation awarded accordingly''. Anti-discrimination law usually looks at the different incidents as distinct cases, not considering possible connections or interrelations. From a societal perspective, though, it is important to recognise these cases because repeated discrimination over time can cause more severe material and immaterial harms to those affected.

Sequential incidents of discrimination can also occur in different steps in a combined process, which produces overlaps with cumulative discrimination (see Example~\ref{ex:hiring_pipeline}). The implications of these two types of discrimination for law and ML are discussed in Section~\ref{sec:cumulativesequential}.

\begin{figure}[!h]
    \centering
    \includegraphics[scale=0.6]{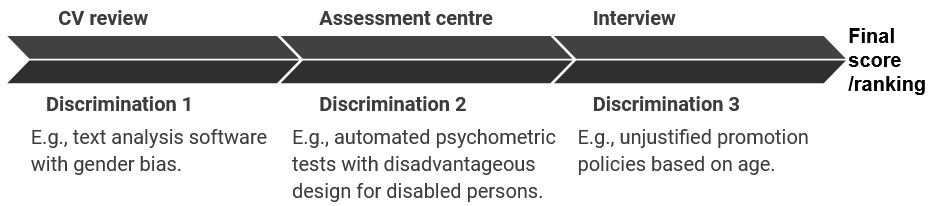}
    \caption{A recruitment process involves several steps with potential for discrimination.}\label{fig:hiring_pipeline}
    \label{fig:hiringpipeline}
\end{figure}

\begin{example}
An example of sequential discrimination in a hiring process is given in Figure~\ref{fig:hiring_pipeline}.
An old woman with disability might suffer discrimination in the first occasion due to her gender, later due to her disability and finally due to her age. Here, the final outcome of the process will be the point at the centre of legal review. 
\label{ex:hiring_pipeline}
\end{example}

\section{Multi-dimensional discrimination in AI systems}
\label{sec:multiDefsML}
We first introduce the typical fairness-aware learning setup and basic concepts (Section~\ref{sec:MLbasics}) and then we survey existing definitions of multi-discrimination in ML, organised into  cumulative (Section~\ref{sec:MLcum}), intersectional (Section~\ref{sec:MLinter}) and sequential discrimination definitions (Section~\ref{sec:MLseq}).


\subsection{Basic concepts and problem formulation}
\label{sec:MLbasics}
We follow the typical fully supervised batch learning fairness-aware setup.
Let $D=(u^{(i)},s^{(i)},y^{(i)})\sim P$ be a dataset of $n$ instances, with each instance drawn as an independent and identically distributed (i.i.d.) sample from 
$P(U\times S \times Y)$, where $U$ is the subspace of \emph{non-protected attributes} (e.g. height, weight, education, etc.), $S$ is the subspace of \emph{protected attributes} (e.g. race, gender, religion, etc.), and $Y$ is the class/target attribute (e.g. loan default). For simplicity, we assume binary classification: $Y\in \{+,-\} $. The non-protected and protected attributes together define the feature space $X=U\times S$, so $x^{(i)}=(u^{(i)},s^{(i)}), 1\leq i\leq n$. Let $S$ consist of $k$ protected attributes: $S=S_1\times S_2 \times \cdots \times S_k$. For simplicity, protected attributes are assumed to be binary: $\forall_{j=1,\cdots,k} S_j\in \{g^j,\overline{g^j}\}$, where $g^j$ and $\overline{g^j}$ represent the \emph{protected group} (e.g. female) and the \emph{non-protected group} (e.g. male), respectively w.r.t. the protected attribute $S_j$ (e.g. gender). 

The intersection of different protected attributes defines the so-called \emph{subgroups}\footnote{We use the term \emph{group} (\emph{subgroup}) if a single (respectively, more than one) protected attribute(s) is used for the definition of the (sub)group.}.
For example, based on the binary protected attributes age, race and gender, eight different subgroups are formed including the subgroups:``young-black-women'' and ``old-white-men''.
The \emph{collection of subgroups} is denoted by  $\mathcal{SG}$ and defines as:
\begin{equation}\label{eq:sg}
  \mathcal{SG}=\{sg_m=s^1 \cap s^2\cap \cdots \cap s^k\mid~s^i\in \{g^i,\overline{g^i}\}, i=1,\cdots k\}\} 
\end{equation} 

Broadly, discrimination for the supervised learning set-up can be expressed in terms of  differences in model performance across different subgroups; these differences can be evaluated w.r.t. one class of interest (typically, the positive class) or w.r.t. both classes. Moreover, model performance can be evaluated in terms of different conditions: just predictions or predictions given the ground truth. We use the generic notation $C$ to denote these extra conditions.

For mono-discrimination, this broad definition can be expressed as differences in expected outcomes of the groups:
\begin{equation}\mathcal{F}_{S_j}\equiv P(\hat{y}\mid g,C)-P(\hat{y}\mid \overline{g},C) - \epsilon
\label{eq:monoDiscr}
\end{equation}
where $\mathcal{F}_{S_j}$ is the \emph{group discrimination} w.r.t. $S_j$, $\hat{y}$ is the predicted outcome, $\epsilon$ is the tolerated discrimination threshold, and 
$C$ refers to the additional conditions w.r.t. class(es) and measure of interest.
For example, Statistical Parity~\cite{dwork2012fairnessParity} only focuses on predictions in the positive class so $C:[\hat{y}=+]$, and  Equation~\ref{eq:monoDiscr} can be re-written as: 
$\mathcal{F}\equiv P(\hat{y}=+\mid g)-P(\hat{y}=+\mid \overline{g})
$.
On the other side, Equal Opportunity~\cite{hardt2016equality} focuses on the correct predictions in the positive class, so $C:$ $[\hat{y}= y$ $\mid y=+]$ and  Equation~\ref{eq:monoDiscr} can be re-written as: 
$\mathcal{F}\equiv P(\hat{y}= y$ $\mid y=+, g)-P(\hat{y}= y$ $\mid y=+, \overline{g})
$.




\subsection{Cumulative discrimination}
\label{sec:MLcum}
Cumulative discrimination is a natural 
extension of mono-discrimination to the multi-discrimination case with the \emph{conceptually isolated groups} defined separately based on each of the protected attributes.
Early works on multi-discrimination~ \cite{agarwal2018reductions,zafar2019fairness} target cumulative discrimination and formulate the problem as solving a \emph{set of fairness constraints}, one for each protected attribute.
Following the generic formulation of Equation~\ref{eq:monoDiscr}, cumulative discrimination over the set of protected attributes $S_1\times\cdots\times S_k$ can be formulated as an operation over a collection of group-specific discrimination: 
\begin{equation}
\label{eq:genMF}
\mathcal{F}_S\equiv \odot (\{\mathcal{F}_{S_1}, \cdots,  \mathcal{F}_{S_k}\})
\end{equation}
where $\mathcal{F}_{S_j}$ is the group discrimination w.r.t. $S_j$, $j=1,\cdots,k$, as defined for the mono-discrimination case (c.f., Eq.~\ref{eq:monoDiscr}), 
 and $\odot()$ is an operator e.g. $max()$, $sum()$ that defines how to ``combine''/``assess'' all-together the multiple group discrimination.
%
More recent works~\cite{yang2020fairness_overlap,roy2022multifair} argue that any fairness notion which aims to find the \emph{maximum} discrimination towards any protected attribute $S_j$ among the set of protected attributes $S$ (i.e. using operator $max()$ for $\odot()$ in Eq~\ref{eq:genMF}), 
is equivalent to the generalised multi-dimensional discrimination formulation of  Equation~\ref{eq:genMF}.




Although in legal practice a separate consideration of grounds in principle allows redress of cumulative discrimination (see Section~\ref{sec:cumulativintersectional}), its application in ML comes with flaws as it targets discrimination in groups defined on single protected attributes but not in subgroups defined based on the intersection of several protected attributes. This drawback was first studied in~\cite{kearns2018gerrymandering} who also termed the drawback \emph{fairness gerrymandering}. In particular, it was shown that a model trained to be individually fair w.r.t. different protected attributes can still discriminate certain subgroups defined based on the intersection of several protected attributes. 

The problem can be elaborated with a hypothetical example.  
During a routine raid by police in some part of the world where drug trafficking is an existing major issue, 
some suspects (say 100) are taken into custody. 
Now assume that based on the protected attribute \textit{gender} the suspected people can be divided into 60:40 \emph{male}:\emph{female}, and based on the protected attribute \textit{race} the distribution is 60:40 {\em black}:{\em white}. Considering both race and gender, 4 subgroups are formed:
(\textit{White Male}, \#20),
(\textit{White Female}, \#20),
(\textit{Black Male}, \#40),
(\textit{Black Female}, \#20). Let us further assume an ML model that is deployed to classify the questioned person as either ``drug trafficker'' or ``innocent''. 
The ML model has been trained to be fair w.r.t. gender and race using the cumulative notion of fairness (cf. Equation~\ref{eq:genMF}) and employing Statistical Parity\footnote{The example would work with other fairness notions as well. We use statistical parity due to its simplicity.} as the underlying fairness notion, i.e. conditioned on the equality of positive predictions between groups defined by gender and between groups defined by race.
\begin{figure}
    \centering
\includegraphics[scale=0.2]{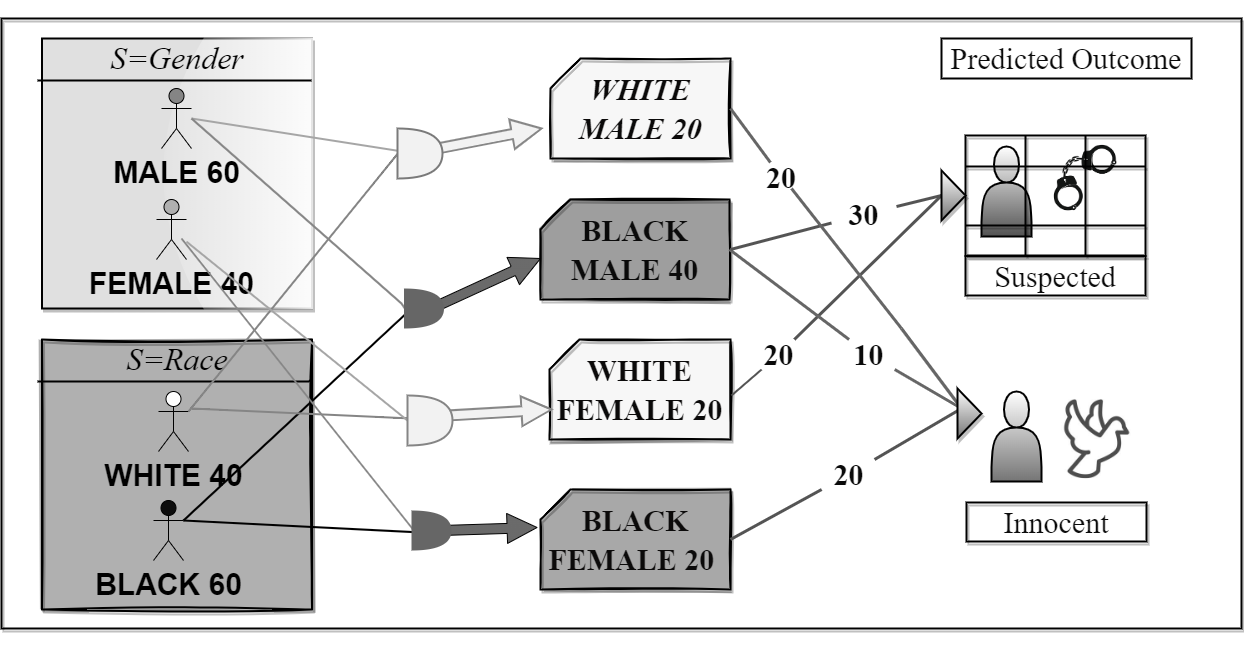}
    \caption{Prediction distribution of a hypothetical drug trafficker detection model for different population (sub)groups.}
    \label{fig:hypoML}
\end{figure}

Figure~\ref{fig:hypoML} illustrates the finer distribution of the population based on gender and race as well as the results of the ML model. As we can see, the model is fair for the different groups: w.r.t. \emph{gender} 50\% of males (30 out of 60) and 50\% of females (20 out of 40) are predicted as suspected \emph{``drug trafficker''}. Likewise w.r.t \emph{race} 50\% of blacks (30 out of 60) and 50\% of whites (20 out of 40) are predicted as suspected \emph{``drug trafficker''}. 
Looking at the subgroups however, the model is unfair e.g. to \textit{white females} (20 out of 20 are predicted as suspected) with 100\% suspect prediction compared to \textit{white males} (20 out of 20 are predicted as innocent) with 0\% suspect prediction.

\subsection{Intersectional discrimination}
\label{sec:MLinter}


Intersectional discrimination looks at the subgroups defined on the intersection of multiple protected attributes (cf. also Section~\ref{sec:intersectionalLaw}).~\cite{pmlr-multicalibration} was the first to study the problem of fairness in finer subgroups, though the limitations of cumulative discrimination and the need to focus on subgroups were clearly outlined in~\cite{kearns2018gerrymandering}. 

A generic definition of intersectional discrimination for a set of protected attributes
$S=\{S_j\}$ 
(adapting from the general definition of Eq.~\ref{eq:genMF}) can be formulated as an operation
over subgroup-specific discrimination:
\begin{equation}\label{eq:genIF}
    \mathcal{F}_{\mathcal{SG}}\equiv \odot(\{ 
   \mathcal{F}_{sg} : {sg} \in \mathcal{SG} \})
\end{equation}
where ${sg} \in \mathcal{SG}$ is a subgroup (cf. Equation~\ref{eq:sg}), $\mathcal{F}_{sg}$ is the discrimination w.r.t. $sg$
and $\odot()$ is an arbitrary operator, e.g. $max()$ that defines how the different subgroup discrimination should be ``combined/interpreted" all-together. 
 
Intersectional discrimination is the most vividly studied multi-dimensional discrimination type in ML literature. Different methods vary mainly w.r.t. how they define  the discrimination for each subgroup, i.e. $\mathcal{F}_{sg}$, therefore, hereafter we focus on this aspect. 
Not all methods propose an explicit ``combination" over the discrimination of the subgroups, but rather try to optimise fairness for each subgroup during discrimination mitigation
~\cite{kearns2019empiricalsubgroup,ma2021subgroup,martinez2021blindsubgroup,subgroupNIPS22}.



\noindent\textbf{Statistical Parity Subgroup Fairness (SPSF)~\cite{kearns2018gerrymandering}:} Kearns et 
al.~\cite{kearns2018gerrymandering} introduced the term ``fairness gerrymandering" to describe the case where a classifier appears to be equitable when considering any protected attribute alone, e.g. only gender or only race, but might be unfair when looking at the intersection of different protected attributes (e.g. black women).
To account for intersectional discrimination they introduced  \emph{Statistical Parity Subgroup Fairness} (\emph{SPSF}), an extension of the \emph{Statistical Parity} (SP)~\cite{dwork2012fairnessParity} definition for mono-discrimination. 

The main idea is that the difference between the acceptance rate (probability of positive prediction) $\mathbf{P}(\hat{y}=+\mid{sg})$ of any subgroup ${sg} \in SG$ from the overall acceptance rate $\mathbf{P}(\hat{y}=+)$ proportional to the relative size of the subgroup $\mathbf{P}({sg})$ in the data, must be smaller than an allowed discrimination threshold $\epsilon$. 
More formally:
\begin{equation}\label{eq:intersecSPSF}
    \begin{split}
      \mathcal{F}_{sg}\equiv \mathbf{P}({sg})\times abs(\mathbf{P}(\hat{y}=+) -\mathbf{P}(\hat{y}=+\mid{sg})) - \epsilon 
    \end{split}
\end{equation}
The threshold $\epsilon\in [0,1]$ quantifies the amount of allowed discrimination towards any subgroup $sg\in \mathcal{SG}$. The relative size of the subgroup $\mathbf{P}({sg})$ allows avoiding fairness overfitting by ignoring discrimination for small fractions of the population (i.e. subgroups which are very small in size, e.g. a singleton subgroup).

A major drawback of this definition is the possibility of high false positive rates 
in order to balance the acceptance rate among the different subgroups; which is a common critique of \emph{SP}~\cite{verma2018fairness,hardt2016equality} and stems from the fact that only predictions but not ground truth are considered.
Further, the method relies upon the subgroup probability $\mathbf{P}(sg)$ estimated from the data and is therefore prone to biased data representations. 
The advantage of this definition lies in scenarios where a subgroup ${sg}$ 
has very few positive instances and comparatively many negative ones. 
In such a case, since the relative size of the subgroup $\mathbf{P}({sg})$ is high, the discrimination in this subgroup w.r.t. the positive class is boosted despite the small number of positive instances.
Such scenarios are highly likely when the number of protected attributes is large. 


\noindent{\textbf{False Positive Subgroup Fairness (FPSF)~\cite{kearns2018gerrymandering}:}} FPSF comprises an extension of the widely used mono-discrimination notion of \emph{Equal Opportunity} (Eq.Opps)~\cite{hardt2016equality} that checks equality of positive predictions between two demographic groups, assuming that people in this group qualify (i.e. the ground truth is positive).
More precisely, it defines subgroup discrimination as the difference between incorrect acceptance (false positive) rate $\mathbf{P}(\hat{y}=+\mid y=-,{sg})$ on a given subgroup ${sg}$ and incorrect acceptance $\mathbf{P}(\hat{y}=+\mid y=-)$ on the entire population (dataset) proportional to the relative size of the negative subgroup $\mathbf{P}(y=-,{sg})$, to be less than a given discrimination threshold $\epsilon$. More formally:
\begin{equation}\label{eq:intersecFPSF}
    \begin{split}
       \mathcal{F}_{sg}\equiv  \mathbf{P}(y=-,{sg})\times abs(\mathbf{P}(\hat{y}=+\mid y=-) - \mathbf{P}(\hat{y}=+\mid y=-,{sg})) - \epsilon 
    \end{split}
\end{equation}

By considering ground truth labels, \emph{FPSF} overcomes the risk
of high false positives rates (see critique on \emph{SPSF}). However, like \emph{SPSF} it relies upon distribution of the subgroups, specifically in the negative (-) class and is therefore prone to biased representations. This has been criticised in
\cite{foulds2020intersectional}, where it is argued that the concept of subgroup fairness,
due to the consideration of the subgroup probability $\mathbf{P}(y=-,{sg})$, is affected by the population size of the subgroup $|{sg}|$. Thus, a discrimination towards a small subgroup ${sg}$ gets unfairly overlooked.

\noindent {\textbf{Differential Fairness (DF)~\cite{foulds2020intersectional}:}}
Foulds et al.~\cite{foulds2020intersectional} criticised the concept of subgroup fairness for its inability to tackle disproportionate distribution of subgroups and proposed \emph{Differential Fairness} (DF) which extends the 80\% rule of the U.S. ``Equal Employment Opportunity Commission"~\cite{equal1978guide} to multiple intersectional subgroups. 
The idea here is to restrict ratios of 
outcome probabilities
between pairs of subgroups under a predetermined fairness
threshold $e^{\epsilon}$. 
More formally:
\begin{equation}\label{eq:intersecDF}
    \begin{split}
        \mathcal{F}_{sg_j,sg_i}\equiv \frac{\max[\mathbf{P}(\hat{y}=c\mid{sg}_j),\mathbf{P}(\hat{y}=y\mid{sg}_i)]}{\min[\mathbf{P}(\hat{y}=c\mid{sg}_j),\mathbf{P}(\hat{y}=y\mid{sg}_i)]}\
        - e^{\epsilon}, ~~c\in\{+,-\} 
    \end{split}
\end{equation}

where $\epsilon$ is an admissible discrimination towards any subgroup, $c$ is the class of interest. The value of $\epsilon$ can be set for different pairs subgroups, which can be determined using various factors such as difference in their data distribution, known historical bias, required economic utility, etc. 

The authors showed that the \textbf{DF} definition closely follows data privacy definitions~\cite{kifer2014pufferfish} and provides provable privacy and fairness guarantees. 
However, the \emph{DF} definition is explicitly designed to extend the 80\% rule~\cite{equal1978guide} between any two subgroups ${sg}_i$ and ${sg}_j$, which identifies
disparate impact in cases where $P(y\mid{sg}_i)/P(y\mid{sg}_j) \leq 0.8$, for a disadvantaged subgroup ${sg}_i$ and best performing subgroup ${sg}_j$. The definition is very closely related to the mono-discrimination definition
{\em Statistical Parity}~\cite{dwork2012fairnessParity} (cf. \ref{sec:MLbasics}), as it focuses only on the predicted output ignoring the ground truth. 

\noindent {\textbf{Worst Case Fairness (WCF)}~\cite{ghosh2021worst}:} A more recent work~\cite{ghosh2021worst} studied the Differential Fairness (cf. Equation~\ref{eq:intersecDF}) and tried to extend the definition which generalises to match all possible mono-discrimination definitions. 
They formulated the discrimination definition as a {\em worst-case} comparison between subgroups under a given condition $C$, where the condition $C$ (cf. Sec.~\ref{sec:MLbasics}) is applied to render the subgroup discrimination definition comparable to a specified mono-discrimination definition~\cite{dwork2012fairnessParity,hardt2016equality,pleiss2017calibration}. Unlike previous works\cite{kearns2018gerrymandering,foulds2020intersectional}, they defined discrimination over the entire collection of subgroups $\mathcal{SG}$ as a min-max ratio of prediction 
probability over any subgroup. Their definition compares worst performing subgroup ($\min\{P(\hat{y}|sg,C)|sg\in \mathcal{SG}\}$) to the best ($\max\{P(\hat{y}|sg,C)|sg\in \mathcal{SG}\}$), formally defined as:
\begin{equation}\label{eq:intersecWCF}
    \mathcal{F}_{\mathcal{SG}}\equiv 1- \frac{\min\{P(\hat{y}|sg,C)|sg\in \mathcal{SG}\}}{\max\{P(\hat{y}|sg,C)|sg\in \mathcal{SG}\}}
\end{equation}
This definition provides the scope to have a good overall evaluation of discrimination, but it fails to provide, when needed, an in-depth information of per-subgroup discrimination independently.   

\noindent \textbf{Discussion:}
Mitigating intersectional discrimination provides the primary advantage of protecting against discrimination of finer sub-populations. However, with many protected attributes, the number of possible subgroups grows exponentially. Even assuming all protected attributes are binary, we get $2^{\mid S \mid}$ subgroups, where $S$ is the set of protected attributes. This gives rise to a problem of data scarcity within the subgroups, which means there exist sub-populations with limited or no data, making it hard to properly (machine) learn the subgroups. 

To better understand the problem, we take a look at the popular ``Adult'' dataset.
The task is to predict whether the income of a person exceeds $50K$/year, with greater than $50K$/year being the positive class.
The dataset contains $\approx 45k$ instances, we consider  \textit{race}, \textit{gender}, and \textit{age} as protected attributes and assume each to be binary~\cite{roy2022multifair}, which gives us 8 subgroups in total.
\begin{figure}
    \centering    \includegraphics[scale=0.4]{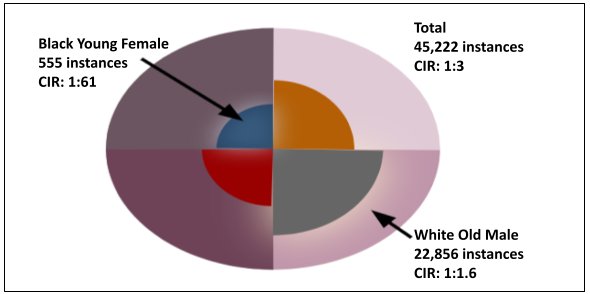}
    \caption{Illustration of the subgroup data scarcity and subgroup class imbalance problems for the Adult dataset.}
    \label{fig:data_scarcity}
\end{figure}
In Fig.\ref{fig:data_scarcity} we highlight the data distribution for some of the subgroups. 
The first observation is that there is a \emph{subgroup scarcity problem} as the subgroups have very different cardinalities. 
For example, the subgroup ``Black Young Female" has only $555~(1.23\%)$ instances, whereas the ``White Old Male" subgroup has almost 41 times more, namely $22,856~(50.54\%)$ instances.
Besides subgroup scarcity, we observe varying positive:negative \emph{class imbalance ratios} (CIR) for every subgroups with higher imbalance for minority subgroups. As a concrete example, the CIR for the underrepresented subgroup ``Black Young Female" is 1:61 (9 positive out of a total 555 instances) which 
more than 38 times higher than majority subgroup ``White Old Male'' with 1:1.6 CIR (8.6k positives out of 22.8k instances). 

\subsection{Sequential discrimination}
\label{sec:MLseq}
Sequential discrimination  by definition requires a sequence of events (see also Section~\ref{sec:sequentialLaw}). The order of the events is very important  
as discrimination at an earlier stage in the sequence can have a larger impact than discrimination at a later stage~\cite{fairpipelines}.
This topic has only recently attracted the interest of the Fair ML community.
Existing works~\cite{pessachshmuelireview} are mainly aimed at long-term (future outcomes) implications~\cite{jabbari2017fairRL,d2020notstaticfair,zhang2020howfairlong,hu2020towards,fairpipelines} of discriminatory outcomes. Also, many of the works are aimed at solving sequential discrimination at an individual level, as it often relies upon feedback from the dynamic/continuous system and how an individual may act after receiving the feedback~\cite{jabbari2017fairRL,d2020notstaticfair,zhang2020howfairlong}. However, staying true to the focus of this work 
we will keep the discussion centred to multi-discrimination within the supervised learning set-up as introduced in Sec.~\ref{sec:MLbasics}. 
However, differently from the batch learning setup of Sec.~\ref{sec:MLbasics}, for sequential discrimination, 
data arrive as a sequence of batches (each batch corresponding to a distinct event, for example, the different steps of the hiring pipeline, see Example~\ref{ex:hiring_pipeline}). More formally, 
$\mathcal{D}=[D_1, D_2, \cdots, D_T]$, where $D_1\cdots D_T$ is in a ordered sequence from $1$ to $T$, and each $D_t\in \mathcal{D}$ consists of $n_t$ instances such that $n_1\leq n_2\leq \cdots \leq n_T$. 
The instances of each event $D_t=(u^{(i)},s^{(i)},y_t^{(i)})\sim P_t$ are drawn as i.i.d samples from the underlying distribution $P_t(U\times S\times Y)$ such that $y_t^{(i)}=1 \implies \forall_{1\leq l \leq t} y_l^{(i)}=1$, i.e. any instance $(i)$ labelled as positive in $D_t$  means a positive label for $(i)$ in all the events observed before $D_t$ (but vice-versa is not true). 
Following \cite{fairpipelines}, for a given protected attribute $S_j\in \{g^j,\overline{g}^jj\}$ and a sequence of observed events $1,\cdots, T$, sequential fairness w.r.t. $S_j$ can be defined as:
\begin{equation}\label{eq:seqmonofair}   
    \mathcal{F}_{S_j}(T)\equiv \frac{P(\hat{y}_T\mid \overline{g},C)}{P(\hat{y}_T\mid g,C)} - \prod_{t=0}^{T-1}\frac{1}{(1+\mathcal{F}_{S_j}(t))}
\end{equation}
where $\mathcal{F}_{S_j}(T)$ is sequential discrimination over the sequence $T$ and $\hat{y}_T$ is the final predicted outcome at the end of the sequence. Notice that the definition in Eq.~\ref{eq:seqmonofair} is recursive and considers in a multiplicative form the discrimination observed from the beginning of the process till previous step $[0,\cdots, (T-1)]$.
Intuitively, the multiplicative property of the definitions is designed to penalise systems for higher discrimination in an earlier step/stage of the decision process. Naturally, at the beginning of the sequence i.e. at $t=0$, the value of $\mathcal{F}_{S_j}(0)$ is not generated by the AI system and is given externally as an input to rectify historical bias known in the society. 
For better understanding, let us consider the 3-stage hiring process example depicted in  Fig~\ref{fig:hiring_pipeline}. Considering gender as the protected attribute, one observes that in CV review step $80\%$ male and $50\%$ applicants get forwarded to the assessment (2nd) step. Now, considering $\mathcal{F}_{Gender}(0)=0$ we get $\mathcal{F}_{Gender}(1)=0.6$. Then, even if in assessment step the system equally accepts both the groups we get $\mathcal{F}_{Gender}(2)=0.37$, and in the interview step to have discrimination (zero) free prediction the system needs to predict with $P('accept'|female)\approx 2P('accept'|male)$, since, $\prod_{t=0}^{2}\frac{1}{1+\mathcal{F}_{Gender}(t)}\approx \frac{1}{2}$. 

The extension of sequential mono-discrimination to a multi-discrimination sequential scenario could be a straightforward practice by defining discrimination $\mathcal{F}_{S_j}(T)$ w.r.t. every protected attribute $S_j$, for $j=1,\cdots,k$, and then combining them using an arbitrary operator $\odot()$ as in Eq.~\ref{eq:genMF}. 
The definition presented in Eq.~\ref{eq:seqmonofair} is also applicable to more than one sub-populations~\cite{fairpipelines}, and thus can be generalised to a multi-discrimination definition on the intersectional subgroups $\mathcal{SG}$ (cf. Eq.~\ref{eq:sg}):
\begin{equation}\label{eq:seqmultifair}
    \mathcal{F}_{sg_k}(T)\equiv \frac{P(\hat{y}_T\mid sg_m,C)}{P(\hat{y}_T\mid sg_k,C)} - \prod_{t=0}^{T-1}\frac{1}{(1+\mathcal{F}_{sg_k}(t))}
\end{equation}
where $\mathcal{F}_{sg_k}(T)$ defines the sequential discrimination on any underprivileged subgroup $sg_k$ (say \textit{black female}) compared to the most privileged subgroup $sg_m$ (say \textit{white male}) in the prediction $\hat{y}_{t-1}$ on the sequence of observed events till $D_{T}$. 

Though no prior work has specifically addressed the sequential multi-discrimination problem, a generalisation of sequential mono-discrimination to the ``multi'' case seems doable. 
The main issue however, is still coming from data scarcity. Most of the work addressing the sequential discrimination problem relies on synthetic data generation~\cite{d2020notstaticfair,zhang2020howfairlong}. Although the ACS-PUMS dataset introduced in~\cite{ding2021retiring} and the Intesa Sanpaolo bank dataset used in~\cite{hu2020towards} possess the temporal property which is important to address a sequential multi-discrimination problem, none of them has the pipeline information which either includes stage-wise label information~\cite{fairpipelines} or change of feature/transitional information due to a decision received in the previous stage~\cite{zhang2020howfairlong}. This signals a need for real-world datasets with temporal and stage-wise information that can be investigated to analyse and develop better sequential multi-fair models.

\section{Discussion}
\label{sec:disc}
After reviewing the fair ML literature w.r.t. the typology from legal scholarship, some aspects bear pointing out. In addition to technical and policy-related challenges, differences in concepts of discrimination and fairness are highlighted by the disciplines' relationship with multi-dimensionality. In the fairly young interdisciplinary discourse on fairness in AI-based decision-making, law and ML have to learn how to incorporate conceptual work from their counterpart.

\subsection{Categorical grounds and attributes in law and ML}
\label{sec:categoricalgrounds}

Fundamental differences can be found in the conceptualisation of protected grounds or attributes, respectively. While the following reflections on categorical grounds and attributes also apply to mono-dimensional fairness, they become particularly apparent once multi-dimensionality is introduced. Perhaps unsurprisingly, the recognition that grounds cannot be neatly separated and defined in isolation often leads to calls for an anti-discrimination law without such categories or with open-ended lists~\cite{lembkeliebscher, essentialismusfalle}.

Rather than innate attributes, intersectional and multi-dimensional concepts of discrimination posit that the ``categories'' named by grounds of discrimination are best understood as categorisations, embedded in a process of social ascription and categorisation. They oppose views which are critically termed ``essentialism'' and entail the assumption that categorisations and questions of identity at the roots of discrimination are (only or predominantly) results of innate and static personal differences. Intersectional analysis emphasises that these bases for discrimination are often context-dependent social phenomena~\cite{makkonen2002multiple}. Structural intersectionality, e.g. discrimination is considered oppression along axes of social power relationships~\cite{fredman2016intersectional, baerbittnergoettsche2010mehrdimensional}. Such a localisation of discrimination in the intersections of oppression is deemed to be central to the framework of intersectionality~\cite{kongintersectional}.

EU law implements a non-essentialist concept of discrimination grounds. The term racial origin, e.g. is used in EU anti-discrimination law to combat \emph{racial~discrimination}, accompanied by an explanation that this does not imply acceptance of theories of separate human races (recital 6 of Directive 2000/43/EC) and the ECJ has applied the law to cases of ``discrimination by association'', implying a non-essentialist reading~\footnote{See ECJ, C-303/06 - \emph{Coleman}, C-83/14 - \emph{CHEZ}; for analysis, see~\cite{xenidis2020_EU_equalitylaw}. Note that at the same time, essentialist interpretations persist, as demonstrated for German jurisprudence by~\cite{essentialismusfalle}.}. 
More specifically, Xenidis\cite{xenidis2020_EU_equalitylaw} argues that EU anti-discrimination law can be read as implementing a non-essentialist dual conception of protected grounds as a recognition of social identities and as a tool to capture social hierarchies. Depending on the context, 
protected grounds can thus capture categorising external ascriptions and stereotypes or group affiliations and identities of the affected persons.

ML, on the other hand, utilises these categories as sensitive \emph{features} or protected \emph{attributes} and often needs to assume their stability. At first glance, ML methodology therefore seems to align more smoothly with an essentialist conception of discrimination. In fair ML, discrimination grounds appear as protected \emph{attributes} or \emph{groups}. Scholarship has emphasised how race is often assumed to be fixed and mono-dimensional, even in work on algorithmic fairness~\cite{hannaetalcriticalrace}. But because an essentialist concept of discrimination grounds can end up reproducing the same group differences and hierarchies that anti-discrimination law and fair ML aim to mitigate, critical scholarship is increasingly calling for a shift of focus in fair ML from protected attributes to structural oppression~\cite{kongintersectional} or social hierarchy~\cite{hoffmannfairnessfails}. Relatedly, work in fair ML research explicitly referencing intersectional theories (such as \cite{kearns2018gerrymandering}) has been criticised for missing the taught by the intersectionality framework by pursuing intersectional fairness only by splitting protected groups further into subgroups~\cite{hoffmannfairnessfails, kongintersectional}. In this regard, Iyola Solanke's concise observation that ``[w]hilst attributes may be innate, stigmas are produced''~\cite{solankeraceandgender}, could be an important lesson for fair ML research.

It is, at the least, a challenge to operationalise the contextuality of anti-discrimination law in ML~\cite{whyfairnesscannot}, and the multi-dimensionality of discrimination, which is only insufficiently reflected in law itself, exacerbates this challenge. Nevertheless, scholarship has already highlighted potential methodological routes beyond fixed attributes for ML~\cite{hannaetalcriticalrace} and for law~\cite{Chege2012, xenidis2020_EU_equalitylaw}. The fairness community extends beyond the design and application of fairness definitions and acknowledges that automated systems are socio-technical systems, e.g. the Web~\cite{berendt2021web}.
A re-conceptualisation of the nature and application of protected attributes can mean looking at the whole socio-technical process of the introduction of AI-based decision-making in a given environment, including awareness of assumptions about – or constructions of – target variables, desired properties and ground truth as well as the selection and construction of protected attributes and respective labels. While some domains particularly require objectivity and stability in class labels (consider, e.g. skin tone differences in cancer screening), in other contexts, labels for protected attributes may be appropriately obtained from affected persons themselves where personal identity or group affiliation matters but from external sources where stereotypes need to be countered. Such a process, informed by social sciences, would go some way towards implementing the dual understanding of protected grounds in anti-discrimination law.


\subsection{Conceptualisation of types of multi-dimensional discrimination}
\label{sec:conceptsmulti}
The terminology regarding different types of discrimination differs; the matter has even been described
as a ``lexical battlefield''~\cite{xenidis2018multiple}. 
Some scholars caution against defining seemingly clear-cut types of multi-
discrimination, calling it a ``dangerously simplifying complication''\cite{baerbittnergoettsche2010mehrdimensional}. Below, we highlight some complexities 
within the typology here applied.

\subsubsection{Cumulative and intersectional fairness}
\label{sec:cumulativintersectional}
As apparent from Figure \ref{fig:boxplot_pums}, cumulative and intersectional discrimination both concern how an individual or group is affected when focusing on \textit{subgroups} defined by two or more discrimination grounds or protected attributes. This common property can be observed in discussions about adequate compensation: For these two types of discrimination, a court finding of discrimination based on only one of the involved grounds may not reflect the discriminated individual's experience\cite{Chege2012} and the extent of the injustice suffered\cite{uccellarimultiple2008}. Some jurisdictions award higher compensation in cases of ``multiple discrimination''~\cite{fredman2016intersectional, chopingermainecomparison} without specifying which types of multi-dimensional discrimination this applies to. Whether higher compensation should be awarded is subject to debate (see~\cite{schiekmultiintorduction} for a brief overview) and higher compensation for both cumulative and intersectional discrimination may be motivated by the argument that individuals in subgroups are more vulnerable in many respects\cite{schiekmultiintorduction}.

However, despite the somewhat blurry line between the two~\cite{baerbittnergoettsche2010mehrdimensional}, in intersectional discrimination intersecting grounds of discrimination are so intertwined that they practically constitute a single criterion applied for differentiation. While this has led scholars to consider intersectional discrimination a case of mono-dimensional discrimination at least in terms of legal doctrine~\cite{weinbergintersektional}, we nevertheless choose to include it under the term multi-dimensional discrimination to emphasise that it results from a combination of multiple \emph{grounds} which are legally often conceived of as separate. Yet, ML approaches to intersectional fairness, by focusing on subgroup fairness, also start out by considering different protected attributes fundamentally intertwined. This makes sense because intersectional discrimination involves the specific challenge of data scarcity for subgroups discussed in \ref{sec:MLinter}. 

The disadvantage of a discriminated subgroup in intersectional discrimination cannot be explained by ``adding up'' disadvantages of two or more groups defined by only one protected ground. Precisely this disadvantageous impact on a subgroup leads to challenges in legal protection that do not pertain to cumulative discrimination: While intersectional and cumulative discrimination both may in some cases be captured by invoking only one of the involved grounds in court or invoking grounds separately~\cite{fredman2016intersectional}, such a strategy is likely to result in complete review for cumulative discrimination (even if potentially understating a claimant's alleged disadvantage), but not for intersectional discrimination. The aforementioned ECJ cases and literature~\cite{crenshaw1989intersectional, Chege2012, lembkeliebscher} have demonstrated that intersectional discrimination tends to elude judicial analysis altogether when the involved grounds are separated. Moreover, differences between cumulative and intersectional discrimination are highly relevant for potential grounds for justification~\cite{weinbergintersektional}.

For ML, on the other hand, the common focus of cumulative and intersectional discrimination on subgroups can play an important role, as Sections~\ref{sec:MLcum} and~\ref{sec:MLinter} demonstrate. Ensuring statistical fairness - by any measure - for subgroups will also do so for groups, thus mitigating intersectional \emph{and} cumulative disparities. Approaches relying only on the notion of cumulative discrimination have the drawbacks highlighted in~\ref{sec:MLcum}, but may be helpful when data scarcity renders an intersectional approach all but practically impossible. 

\subsubsection{Cumulative and sequential fairness} \label{sec:cumulativesequential}

In sequential discrimination each of the multiple incidents of discrimination in a temporal order may involve any other type of discrimination (mono-dimensional, cumulative or intersectional). Importantly, there is a potential overlap with cumulative discrimination.
Most legal tools against discrimination, such as liability, are retrospective and require a ``discriminatory treatment'' or ``particular disadvantage'' in an area covered by anti-discrimination law (employment and, to a lesser degree, supply of goods and services). Legal redress is thus often restricted to a treatment, criterion, provision or practice with a direct and tangible \emph{economic impact} on the affected person~\footnote{This applies especially to jurisdictions relying on individual claimants to enforce anti-discrimination law. In C-54/07 - \emph{Feryn}, par. 21-28, however, the ECJ ruled (notably following a type of class action under Dutch law) that an identifiable affected individual was not required for a finding of discrimination if an employer publicly declares their intention to discriminate in hiring.}.
In hiring, e.g. the rejection of a candidate will typically be at the centre of analysis. But applicants usually undergo several steps in a recruitment process, e.g. CV review, assessment centre tasks or psychometric measurement and interviews as shown in Figure~\ref{fig:hiringpipeline}. Each of these practices bears potential for discrimination~\footnote{For examples of potential biases, see https://www.reuters.com/article/us-amazon-com-jobs-automation-insight-idUSKCN1MK08G for automated CV review and https://www.technologyreview.com/2021/02/11/1017955/auditors-testing-ai-hiring-algorithms-bias-big-questions-remain/ for automated assessment center applications.}. If different steps disadvantage a candidate based on different discrimination grounds, the final decision will often present itself as cumulative discrimination.

In ML, however, cases involving cumulative discrimination from a legal perspective may be addressed as a special type of sequential discrimination from an \emph{engineering perspective}: the process can be segmented and each step addressed according to the sequence. Such use-case are sometimes referred to as \emph{fair pipelines}~\cite{pessachshmuelireview}. Research has shown that under certain conditions interventions at one stage can even propagate through the whole process~\cite{fairpipelines}. Due to the presence of a well-defined task and outcome (e.g. fill an open position) such cases offer more concrete options for interventions for fairness than the ``typical'' sequential discrimination from legal scholarship (cf. Fig. \ref{fig:sequential}), where the effects of discrimination at one decision point on subsequent decisions are hard to determine. Sequential scenarios are well-known and highly important for ML: learning a model is typically the result of a multi-step process, from data selection to pre-processing, cleaning, model selection and evaluation. Bias can penetrate in each of these steps; e.g. w.r.t. dataset selection a strong bias towards certain demographics has been shown in visual datasets~\cite{fabbrizzi2022survey} or, w.r.t. pre-processing, it has been shown that the encoding method for categorical protected attributes can lead to biased models~\cite{mougan2022fairness}. Addressing such cases as sequential discrimination seems a promising route for fair ML.

\section{Outlook}
\label{sec:outlook}
We have introduced the multi-dimensionality of discrimination, taken from legal scholarship, as a not yet fully explored foundational problem of fairness in AI-based decision-making. Looking more closely at different types of discrimination, we can learn how to better address them in decision-making processes. Our review of ML research has shown that the field has begun to address some of the issues raised by multi-dimensionality, predominantly focussing on intersectionally fair algorithms. We have also pointed out obstacles to a common understanding of protected grounds and attributes, and highlighted concerns that essentialist approaches insufficiently reflect multi-dimensionality. These findings raise questions that legal and computer science scholarship have just begun to explore. We can only point to a few of these:

\noindent \textbf{Sources and definition of protected attributes:} Research reflecting on protected attributes is needed to fully appreciate the multi-dimensionality of discrimination. This begins with the question \emph{which} protected attributes to use in a given context: Should these be limited to the grounds covered by anti-discrimination law or address further disadvantages? This choice is even more challenging when multiple jurisdictions are involved~\cite{Legal-technical}. More work is needed on the expansion of fairness frameworks to more protected grounds. Subsequently, protected attributes need to be defined and data labelled accordingly.
This process needs to be informed by other disciplines, especially social sciences and law, where cases are available.

\noindent \textbf{Fairness trade-offs between (sub-)groups:} With an increasing number of protected attributes, including subgroups, it becomes more likely that increasing fairness for one attribute limits or decreases fairness for others. Beyond a balancing of rights~\cite{Gentgenbarg2022}, the law does not provide clear guidance for such scenarios. A factual ``hierarchy'' of discrimination grounds exists w.r.t. scope and strictness of protection~\cite{howard2006hierarchy, holzleithnerdisentangling}, but it seems unlikely that it was intended by the lawmakers to be applied to direct trade-offs. Doing so could incite a ``battle of oppressions''~\cite{holzleithnerdisentangling} on who is more deserving of protection. While this is a known issue in anti-discrimination law, it may become more pressing when directly laid open by the seeming precision of statistical fairness measures. Future work should address these questions.

\noindent \textbf{Data scarcity for intersectional subgroups:} On a practical level, methods to assess fairness under the condition of data scarcity (see, e.g. \cite{vealebinnssensitivedata} for tentative ideas) are important to detect and address cumulative and, especially, intersectional forms of discrimination as the collection of more data on protected attributes, including subgroups, meets various challenges\cite{whatwecantmeasure}. In the legal domain, much comes down to the question of enabling the collection and use of strictly regulated sensitive data (cf. art. 9 GDPR) for fair ML while ensuring data privacy~\cite{vanbekkum_sensitivedata}. Art. 10 (5) of the planned EU AI Act~\footnote{Proposal for a Regulation of the European Parliament and of the Council laying down harmonised Rules on Artificial Intelligence (Artificial Intelligence Act) and amending certain Union legislative Acts, COM(2021) 206 final, 21st April 2021.}, allowing the use of sensitive personal data for bias monitoring, detection and correction may be a step towards a new balance. Data scarcity could be also addressed in other ways, for example by generating synthetic data of desired characteristics~\cite{dos2022generating}. However, describing the characteristics of the targeted subgroups is not an easy task and might introduce subgroup biases and prejudices and lead to both allocative and representational harms.
Even if data is available, statistical tests applied in ECJ jurisprudence are not always suitable for identifying discrimination of small minority groups~\cite{whyfairnesscannot}, which is a particular problem for intersectional discrimination. Thus, work on suitable statistical tests for discrimination under the law is needed.

\noindent \textbf{Sequential scenarios:}
ML models are the result of complex pipelines with several components and decisions affecting the resulting models. As bias and discrimination w.r.t. a single or more protected attributes can arise at any stage of the pipeline, it is important to take into account the discriminatory effects of these components in the overall pipeline and address them holistically rather than in isolation in order to improve the overall utility of the model.


\bibliographystyle{ACM-Reference-Format}
\bibliography{literature}


\end{document}